\pdfoutput=1
\documentclass[11pt]{article}
\usepackage{pdt_report}
\usepackage{algorithm}
\usepackage{algorithmic}
\usepackage{amsmath}
\usepackage{amssymb}
\usepackage{amsthm}
\usepackage[toc,page]{appendix}
\usepackage{array}
\usepackage{booktabs} 
\usepackage{bm}
\usepackage{cancel}
\usepackage{lipsum}
\usepackage{mathtools}
\usepackage[framemethod=default]{mdframed}
\usepackage{multicol}
\usepackage[most]{tcolorbox}
\usepackage{graphicx}
\usepackage{subcaption}
\usepackage{titletoc}
\usepackage{tikz}
\usepackage{wrapfig}
\usepackage{xcolor}
\usepackage[textsize=tiny]{todonotes}

\usetikzlibrary{bayesnet}

\theoremstyle{plain}

\theoremstyle{definition}

\theoremstyle{remark}

\definecolor{softteal}{RGB}{70, 190, 180} 

\def\eqref#1{equation~\ref{#1}}

\def\1{\bm{1}}

\DeclareMathAlphabet{\mathsfit}{\encodingdefault}{\sfdefault}{m}{sl}
\SetMathAlphabet{\mathsfit}{bold}{\encodingdefault}{\sfdefault}{bx}{n}

\usepackage{makecell}

\definecolor{boilermakergold}{HTML}{cfb991}
\definecolor{steel}{HTML}{555960}
\definecolor{coolgray}{HTML}{6f727b}
\usepackage[pagebackref,breaklinks,colorlinks,allcolors=coolgray]{hyperref}
\usepackage{placeins}
\usepackage{cleveref}
\crefname{equation}{Eq.}{Eqs.}
\Crefname{equation}{Eq.}{Eqs.}
\crefformat{equation}{Eq.~(#2#1#3)}
\Crefformat{equation}{Eq.~(#2#1#3)}
\crefname{section}{Sec.}{Secs.}
\crefname{subsection}{Sec.}{Secs.}
\crefname{figure}{Fig.}{Figs.}
\Crefname{figure}{Fig.}{Figs.}
\crefname{table}{Tab.}{Tabs.}
\crefname{appendix}{Appendix}{Appendices}
\Crefname{appendix}{Appendix}{Appendices}

\newcommand{\seedcell}[2]{\ensuremath{\underset{\pm #2}{#1}}}

\title{%
  How Can Driving World Models Do Counterfactual Prediction?
}

\author{%
  Jiaru~Zhang$^{1}$\thanks{Corresponding author: Jiaru Zhang.}~,
  Can~Cui$^{2}$, Yi~Xu$^{2}$, Xin~Ye$^{2}$, Ruqi~Zhang$^{1}$, Ziran~Wang$^{1}$\\
  $^{1}$Purdue University \quad $^{2}$Bosch Center for Artificial Intelligence\\
  \texttt{\{jiaru, ruqiz, ziran\}@purdue.edu}
}

\begin{document}

\maketitle

\begin{abstract}
Driving world models are often interpreted as counterfactual simulators for observed driving episodes: given a factual driving log, they are asked what would have happened under an alternative ego action. In this paper, we identify a fundamental mismatch between this goal and direct action-conditioned prediction. The direct prediction uses the shared history and the alternative action but not the factual continuation observed after that history. It can therefore generate a plausible future without preserving what actually happened in this episode. We formalize this gap using the causal recipe of abduction, action, and prediction and study it in a setting with a short time horizon, where the alternative ego action does not alter how surrounding agents evolve. To make the gap measurable, we construct a controlled simulation benchmark with factual outcomes and matched counterfactual outcomes. Across two representative world models, direct predictions fail to match the counterfactual ground truth, supporting our analysis. As a constructive check of this analysis, we introduce a deliberately simple, training-free pipeline that moves observed evidence into the counterfactual view and lets the frozen model complete what remains unknown. Even this simple construction raises the overall recovered fraction substantially and reduces perceptual distance to the matched counterfactual on both models. We hope this work draws attention to this gap and motivates better counterfactual prediction methods for driving world models.
\end{abstract}

\section{Introduction}
\begin{quote}
{\footnotesize
``\ldots{}What might have been is an abstraction / Remaining a perpetual possibility /
Only in a world of speculation\ldots''}\\[3pt]
\mbox{}\hfill{\scriptsize -- T.S. Eliot, ``Burnt Norton''}
\end{quote}

\noindent \emph{What might have been} is the defining object of counterfactual prediction.
This problem arises throughout autonomous driving, where we often want to know the result of an alternative action.
For example, suppose we have a short factual driving log of what actually happened, which shows a car emerging from a side street or another vehicle cutting in. After
observing the episode, we ask what the camera \emph{would have} recorded had
the ego followed a different trajectory, accelerating or braking to a
stop.
In this setting, the ego's alternative action changes
only its viewpoint and direct consequences. A faithful answer must therefore remain tied to the recorded episode:
the same car emerges or cuts in, while what follows from the ego's changed motion may differ.
This is what separates a counterfactual from an ordinary prediction, which may
return any plausible continuation consistent with the history.

Currently, driving world models are widely considered capable of counterfactual prediction.
For example, Vista
claims a
``\textit{counterfactual reasoning ability}'' to ``\textit{predict the counterfactual
consequences caused by abnormal actions}'' \citep{vista}. Drive-WM states that ``\textit{our
Drive-WM can generate counterfactual events}'' \citep{drivewm}. Industrial world models have likewise made related counterfactual claims \citep{waymowm, genie3}.
The standard paradigm for counterfactual prediction is straightforward. The models receive an alternative
action, typically represented by a target trajectory, and generate the
corresponding future.
The underlying assumption is that conditioning the model on a counterfactual action is sufficient to yield a valid counterfactual prediction \citep{vista, drivewm}.
We refer to the resulting output as the direct action-conditioned
prediction, or simply the direct prediction.

\begin{figure}[t]
\centering
\includegraphics[width=\linewidth]{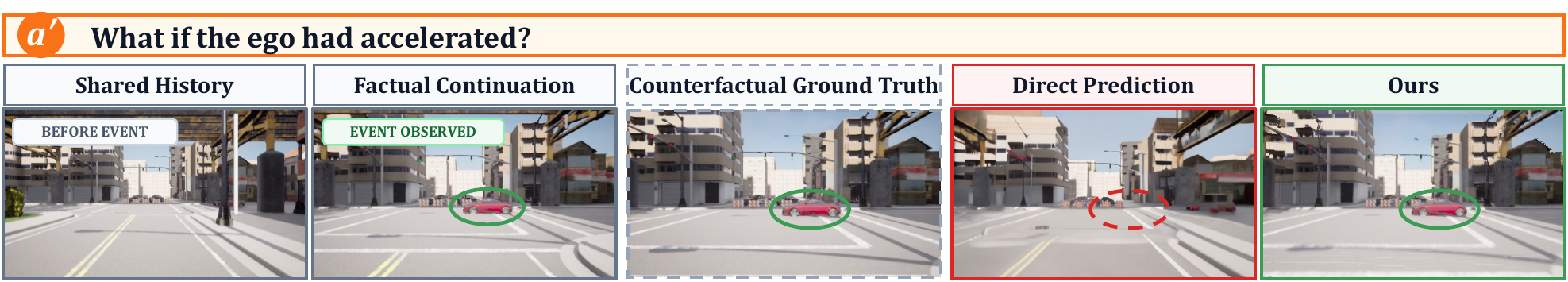}
\caption{Counterfactual prediction for one observed episode. The shared history shows
the ego approaching an intersection. In the factual continuation, a red car emerges
from a side street as the ego follows its recorded trajectory. We ask what the camera would have recorded had the ego instead accelerated
along the target trajectory $a'$ shown in the header. The
counterfactual ground truth shows that the car would still have emerged. The
direct prediction, which uses the shared history and $a'$,
fails to preserve this event, whereas Ours additionally uses the factual
continuation and recovers the car near its counterfactual location.}
\label{fig:teaser}
\end{figure}

However, a mismatch lies in the information each prediction uses. The direct action-conditioned prediction only conditions on the shared history and target
trajectory, while the requested counterfactual must also account for the factual continuation.
In the case above, the shared history alone does not reveal
whether or when a vehicle later emerged from the side street or cut in. Without that evidence, the model has no basis for preserving the particular event in that episode and can return a fluent video that is consistent with the action but omits the specific factual event.
\cref{fig:teaser} illustrates this failure.
The shared history shows the ego approaching an intersection, while a later frame from the factual continuation shows a car emerging from a side street. Given the shared history and ego
acceleration, the direct prediction does not preserve this emergence, even
though running the same world again under that action shows that it would still
have occurred.

In this paper, we first give this failure a precise causal reading.
It reflects a confusion between the direct prediction and the counterfactual prediction, a distinction made precise by Pearl's ladder of causation \citep{pearlcausality, bookofwhy}.
The ladder separates three questions. Rung~1 is the conditional prediction from passive observation, rung~2 is the effect of an intervention in general, and rung~3 is the counterfactual prediction, i.e., what would have happened under a different action in a specific episode that has already been observed. 
A genuine counterfactual prediction, which answers what would have happened in this very world had we acted differently, requires \emph{abduction}, which is the process of using the observed factual continuation to infer the underlying state of that specific world.
The direct prediction fails to do this. 
Because current driving world models are queried with an alternative action using only the shared history, ignoring the factual continuation, their output is a general future under that action rather than a precise counterfactual prediction for this particular episode.

To make the distinction measurable, we need the counterfactual ground truth, i.e., the video that would have been recorded under the alternative action in the same world.
However, real driving can never provide it, because each episode is observed under exactly one action and the future under any alternative action will never be recorded.
As an alternative, we turn to the CARLA simulator \citep{carla}, where the same world can be simulated multiple times under different actions, producing an exact reference in our controlled setting.
Using this simulator, we build a benchmark for counterfactual prediction in driving scenarios. 
Each case provides a factual driving log, including the full RGB video and its synchronized executed ego trajectory, together with a target trajectory that defines the counterfactual query.
Running the same world with the alternative action yields the counterfactual ground truth. 
This ground truth hence enables an evaluation protocol that can quantitatively compare the performance of different prediction methods.

Using this evaluation protocol, we quantitatively demonstrate that direct predictions from both diffusion-based and autoregressive driving world models struggle with counterfactual prediction. To address this, we propose a deliberately simple framework grounded in our causal formulation.
Concretely, it first transports evidence from the factual continuation into the counterfactual camera view.
Then, it uses world models to complete what that evidence does not determine.
The framework requires no training and leaves the model weights unchanged.
Experiments on our benchmark show that it recovers much of the event signal lost by direct prediction and substantially improves visual fidelity across both model families.
In summary, our contributions include:
\begin{itemize}\itemsep2pt
\item We identify and causally analyze the gap between direct action-conditioned prediction and true counterfactual prediction, attributing this failure to the omission of evidence from the factual continuation.
\item We construct a controlled CARLA benchmark with simulated counterfactual ground truths, making this gap quantitatively measurable.
\item We provide a simple training-free construction that supplies the missing evidence to frozen world models. On our benchmark it recovers much of the event signal lost by direct prediction, which confirms the diagnosis and gives future methods a reference point.
\end{itemize}

\section{Related Work}

\subsection{Driving World Models and Their Counterfactual Claims}

Driving world models learn to predict future driving observations or scene states from recorded context. Many video-based models additionally support control through ego actions, trajectories, or commands. Diffusion-based examples include DriveDreamer \citep{drivedreamer}, which generates videos from an initial frame and structured traffic information; Drive-WM \citep{drivewm}, which generates controllable multiview futures and uses them for planning; and Vista \citep{vista}, which supports high-fidelity generation and versatile action control. Autoregressive examples include GAIA-1 \citep{gaia1}, which predicts image tokens from video, text, and action context; OccWorld \citep{occworld}, which tokenizes 3D occupancy and forecasts future scene and ego states; and DrivingWorld \citep{drivingworld}, which employs a pose-conditioned video GPT.

Many of these models claim counterfactual prediction capabilities. Vista and Drive-WM present videos generated under abnormal or alternative maneuvers as evidence of counterfactual reasoning. 
Similarly, industrial world models make related claims. Waymo demonstrates counterfactual driving by simulating a past recorded drive under an alternative route \citep{waymowm}. Google DeepMind's general-purpose Genie 3 supports promptable world events for counterfactual scenarios \citep{genie3}.
Related work also studies vision-language counterfactual reasoning in OmniDrive \citep{omnidrive}, controllable safety-critical traffic generation in CCDiff \citep{ccdiff}, and action-controlled video simulation and reward estimation in ReSim \citep{resim}.

\subsection{Counterfactual and Causal Prediction}

In causality, counterfactual prediction has a precise definition. Pearl's hierarchy
organizes the questions a learner can answer into observational prediction,
interventional effects, and counterfactuals over a specific realized
episode \citep{pearlcausality, bookofwhy}, and the hierarchy is strict, in
that higher rungs are in general not identifiable from information at lower
rungs alone \citep{pch}. In structural causal models the
counterfactual has an exact recipe: abduction of the exogenous state from
the observed outcome, replacement of the action, and prediction through
the same mechanism \citep{pearlcausality}. The tradition of potential outcomes
formalizes the same object as the outcome a unit would have exhibited under
a different treatment, and reality reveals only one of the potential
outcomes per unit \citep{holland1986}. When the available data and causal assumptions do not uniquely identify a counterfactual distribution, classical work characterizes the resulting bounds
\citep{balke1994,manski2003}. 
A recent position paper argues that causal considerations are essential for foundation world models in embodied AI and calls for concrete counterfactual tasks and evaluation metrics \citep{causalwm}.

\section{Problem Setup and Counterfactual Analysis}
\label{sec:analysis}

\subsection{Problem Setup}
\label{subsec:setup}
We formalize a counterfactual prediction task based on a short, fully recorded factual driving log. 
This log contains an RGB video from the front camera synchronized with the executed ego motion. 
It records a particular driving episode in which the ego follows a path while surrounding agents may cut in, cross the road, or brake. 
We then pose the counterfactual query, asking what the ego camera would have recorded had the ego executed a different action, such as early braking or acceleration. Questions of this form arise when vehicle logs are examined after the fact for incident analysis, safety auditing, or liability assessment. In these settings, the complete factual log exists by definition, and the value of the answer lies in its being about this realized episode rather than an arbitrary plausible continuation.

\noindent\textbf{Inputs.} At query time, we are given a factual driving log
$(F,a_{\mathrm{obs}})$, comprising the full video $F$ and its synchronized executed ego trajectory $a_{\mathrm{obs}}$, represented by the vehicle's position and heading at each frame.
The shared history $H$ is the initial synchronized prefix of the
factual log, including the video and ego motion, which is common to both the executed and target trajectories.
We write $F^{+}$ for the RGB frames in $F$ after the shared history interval, namely the factual continuation under the executed trajectory. 
Although these frames occur later on the episode timeline, they are observed evidence when the query is posed. 
The query additionally
specifies the alternative action $a'$, represented by its target position and
heading at each frame. 

\noindent\textbf{Expected output.} The desired output is the video that would have followed $a'$ in
\emph{this same world}. In our setting, this means preserving what actually happened and changing only what the new action directly affects.

\noindent\textbf{Scope.} We study counterfactuals over a short time horizon where the alternative ego action alters the camera viewpoint while the surrounding environment follows predetermined behaviors. This open-loop setting fits queries about the second or so after the ego action changes, since driver perception and reaction times are themselves on the order of a second \citep{green2000}, leaving surrounding agents little time to respond. It makes the question empirically checkable, since a world whose other agents
follow predetermined behaviors can be replayed under the alternative action to
record the outcome that would have followed.

\subsection{What Causal Theory Prescribes}
\label{subsec:estimand}

Our target counterfactual is the outcome under the alternative action $a'$ for the same realized world, where the other agents move as they did.
In causal terms, this is a counterfactual at rung~3 conditioned on the outcome
\citep{pch}. 
It conditions on the factual continuation $F^{+}$ in addition to $H$, thereby updating what is known about the realized episode. Direct prediction instead uses $H$ and $a'$ alone.
Letting $Y$ denote the video that follows the history under a given action and $Y_{a'}$ its value under $a'$, the two are
\begin{equation}
\underbrace{p\big(Y_{a'} \mid H,\, F^{+}\big)}_{\text{counterfactual prediction}}
\quad\text{vs.}\quad
\underbrace{p\big(Y \mid H,\, a'\big)}_{\text{direct prediction}}.
\label{eq:estimand}
\end{equation}
For simplicity, we leave the original action $a_{\mathrm{obs}}$
implicit.

Causal theory also prescribes, in full generality, how the target should be computed
\citep{pearlcausality}. Letting $G$ denote the mechanism taking the world $w$ and an action to an outcome, the
observation itself arose as $F^{+}=G(w,a_{\mathrm{obs}})$. We have
\begin{equation}
\underbrace{w \sim p(w \mid H, F^{+})}_{\text{abduction}}
\;\longrightarrow\;
\underbrace{\vphantom{p(\,)}a'}_{\text{action}}
\;\longrightarrow\;
\underbrace{\vphantom{p(\,)}Y_{a'}=G(w,a')}_{\text{prediction}}.
\label{eq:recipe}
\end{equation}
First comes \emph{abduction}, which uses the shared history and factual
continuation to infer the state of the world that was realized. Next, the \emph{action} step applies
the alternative ego action. Finally, \emph{prediction} propagates the recovered
state through the mechanism under the new action. 

\subsection{Why Direct Prediction Is Insufficient}
\label{subsec:standard-practice}
In practice, world models are trained on (history, action, future) triples to model $p(Y\mid H,a)$. Given $H$ and $a'$, the direct procedure generates $B \sim p(Y\mid H,a')$, which we call the \emph{direct prediction}. Such outputs are commonly regarded as counterfactual predictions \citep{vista,drivewm}.
As $a'$ is specified by the query rather than observed as evidence, it
does not update the posterior over $w$ beyond $H$. 
Therefore, only when the logged action
carries no information about the world beyond the history does conditioning on
$a'$ coincide with intervening, and the right side of \cref{eq:estimand}
equals the interventional prediction $p(Y \mid H, \mathrm{do}(a'))$ at
rung~2, where $\mathrm{do}(a')$ sets the ego action to $a'$. However, the remaining gap to the counterfactual lies in the conditioning
evidence alone, since rung~3 additionally conditions on the factual outcome
$F^{+}$.
Written as mixtures over
the world, the two sides
of \cref{eq:estimand} are
\begin{align}
p(Y \mid H, a') &= \int p\big(Y \mid w, a'\big)\, p(w \mid H)\, dw,
\label{eq:direct}\\
p(Y_{a'} \mid H, F^{+}) &= \int p\big(Y \mid w, a'\big)\, p(w \mid H, F^{+})\, dw.
\label{eq:cf}
\end{align}
The direct prediction, \cref{eq:direct}, infers the world from $H$ alone. It therefore
mixes over $p(w\mid H)$, whereas the counterfactual uses the posterior for this
episode, $p(w\mid H,F^{+})$. Whenever $F^{+}$ carries outcome information
absent from $H$, these two distributions differ.

\section{Methodology}
\label{sec:method}

\subsection{Overview}
\label{subsec:twohalves}

\noindent\textbf{Problem decomposition.}
As shown in \cref{eq:direct,eq:cf}, the factual continuation $F^{+}$ updates what is known about the realized world. 
Therefore, to instantiate the abduction step of \cref{eq:recipe}, we ask which parts of the counterfactual view this evidence determines. The answer splits into two.
One part of the counterfactual view shows surfaces that $F$ also observes,
where, given the sensing setup above, the counterfactual differs only in
where the camera is. On that part the posterior
$p(w \mid H, F^{+})$ concentrates on the observed surfaces, and the outcome
is determined up to the error of recovering their geometry from monocular
video. 
Other parts remain unseen in $F$, such as regions behind occluders or beyond
the field of view. Since two worlds can produce the same observations in $F$ and still
differ in unobserved regions, the posterior retains uncertainty in these regions and every
completion consistent with the evidence is admissible.

\noindent\textbf{Our approach.} The split above assigns the work. Where the world is pinned by evidence, the task is transport that carries the observations into the new viewpoint. Geometric methods based on depth reprojection or novel view synthesis \citep{3dphoto, 3dgs, streetgaussians, gcd, freevs} can do this. The remaining regions require a prior that chooses whether the area behind a bus contains road, curb, or another car, and a driving world model supplies it through its history-and-action interface, which yields the mixture over $p(w \mid H)$ in \cref{eq:direct}. Our method bridges the two. Geometry transports the observed evidence where it determines the answer, the frozen world model completes the scene wherever only a prior can, and the final Combine stage restores the transported evidence in the output.

\subsection{Pipeline}

\begin{figure}[t]
\centering
\includegraphics[width=\textwidth]{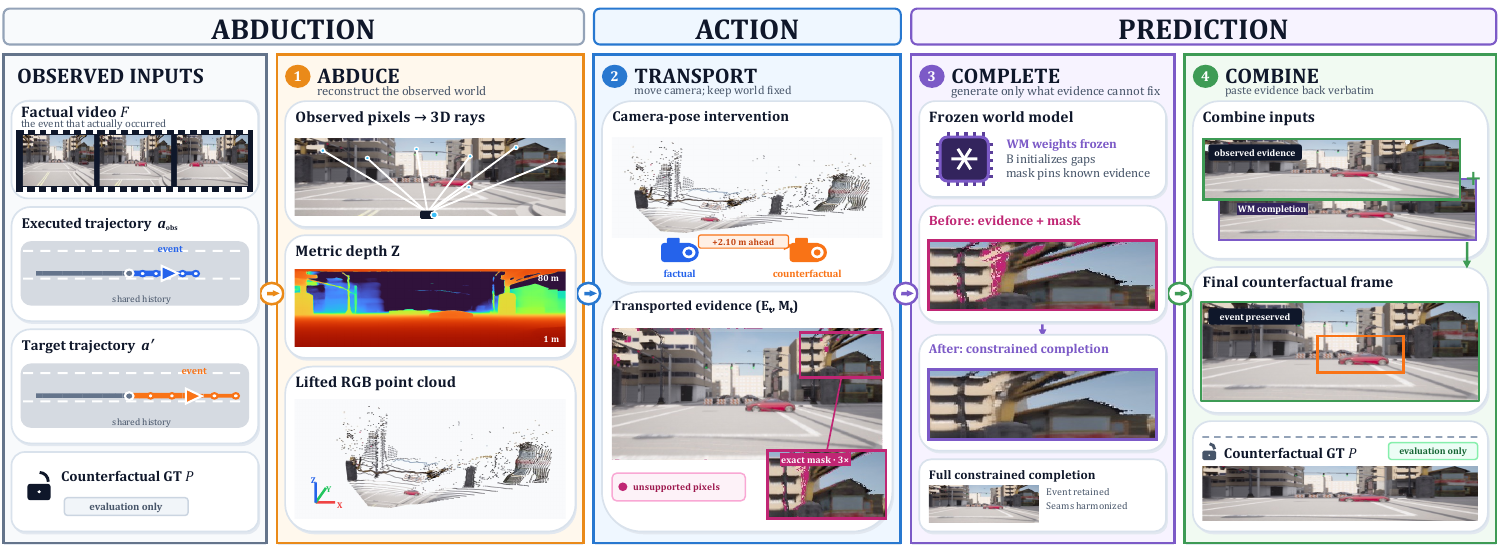}
\caption{Overview of our method. The factual driving log comprises RGB video
$F$ and its executed ego trajectory $a_{\mathrm{obs}}$, while the target trajectory
$a'$ specifies the counterfactual action. The four stages instantiate the
causal recipe of abduction, action, and prediction. (1) \textbf{Abduce} recovers the
observed part of the realized world from the factual log. (2)
\textbf{Transport} applies the target action by moving the camera from
$a_{\mathrm{obs}}$ to $a'$ while holding the world fixed, yielding factual
evidence $E_t$ and a support mask $M_t$ in the counterfactual view for each
frame $t$ of the prediction window. (3--4)
\textbf{Complete} and \textbf{Combine} implement prediction: the frozen world
model generates the unsupported regions, while the Combine stage restores the
transported evidence in the output. The counterfactual ground truth $P$, the replay of the same world under $a'$
defined in \cref{sec:benchmark}, serves as an evaluation reference.}
\label{fig:pipeline}
\end{figure}

As shown in \cref{fig:pipeline}, our method follows the three steps from
\cref{eq:recipe} with four stages. \emph{Abduce} implements the observable part of abduction,
\emph{Transport} implements the action step, and \emph{Complete} and
\emph{Combine} together implement the prediction step.
Together, these stages implement the split in
\cref{subsec:twohalves}.

\noindent\textbf{Abduce.}
A frozen depth model \citep{depthanything} estimates the relative distance of
each pixel from the camera in every frame of $F^{+}$. We use the visible road
and the approximate height of the camera above it to convert these values to
distances in meters. The fixed camera model then maps each observed pixel to a colored 3D
point, forming an RGB point cloud of the observed scene.

\noindent\textbf{Transport.}
Because the camera is rigidly mounted, the executed trajectory
$a_{\mathrm{obs}}$ and target trajectory $a'$ determine,
at each time step, the relative pose between the camera that did film
the scene and the camera that would have filmed it. Then, forward splatting with a depth
buffer reprojects the lifted points into the counterfactual view, producing a warped evidence image
$E_t$ and a support mask $M_t$:
$(E_t,M_t)=\mathrm{splat}\big(\mathrm{lift}(F^{+};a_{\mathrm{obs}}),\,
\mathrm{cam}_t(a')\big)$.
The mask marks the \emph{supported region}, the pixels of the counterfactual
frame whose corresponding 3D point is visible in $F^{+}$. 
Within this region, each pixel takes its value from the factual pixel observing
the same surface point. The time-aligned factual frame is the primary donor because it shows moving
agents at the correct time. To fill residual holes, we further admit projections
from other frames of $F^{+}$, favoring pixels whose projections agree across
frames.
We refer to this refinement as filling from multiple frames (MF).

\noindent\textbf{Complete.}
The frozen world model generates the unsupported regions under the target
trajectory $a'$ while preserving the transported evidence. We first construct
an input video that uses $E_t$ where $M_t=1$ and the corresponding frame of the
direct prediction $B$ everywhere else. For the diffusion model, sampling starts
midway through the denoising process from a noisy encoding of this input video
\citep{sdedit}. After every denoising step $i$, the evidence region is restored
at the corresponding noise level \citep{repaint},
$x \leftarrow M\odot\big(z_{E}+\sigma_i\,\varepsilon\big)+(1-M)\odot x$,
where $x$ is the current video representation, $z_E$ is the encoded input
video, $M$ holds the support masks $M_t$ resampled to the resolution of
$z_E$, $\sigma_i$ is the current noise level, and $\varepsilon$ is random
noise. Each cell of $M$ stores the fraction of its pixels covered by evidence
(\cref{app:method}). For the autoregressive
VQ model, tokens covered by transported evidence stay fixed to the input-video
tokens, while the model generates the remaining tokens normally.

\noindent\textbf{Combine.}
Encoding and decoding through the world model can blur transported pixels. The
Combine stage restores the reliable transported pixels in the completed frame
and blends their boundary smoothly, giving the output frame $\hat{Y}_t$,
$\hat{Y}_t=\alpha_t\odot E_t+(1-\alpha_t)\odot \mathrm{cc}(C_t)$,
where $C_t$ is the completed frame. The weight $\alpha_t$ is $1$ inside the
reliable part of the support mask and gradually decreases to $0$ near its
boundary. The map $\mathrm{cc}$ adjusts the colors of $C_t$ to match the
transported evidence; the adjustment stays within a small range and changes
smoothly across frames.

Our method runs entirely at inference time with all pretrained networks
frozen. Its case-specific inputs are the factual driving log
$(F,a_{\mathrm{obs}})$ and target trajectory $a'$. The camera setup is shared
across cases. More implementation
details are provided in \cref{app:method}.

\section{Benchmark and Metrics}
\label{sec:benchmark}

\subsection{The Benchmark}

\noindent\textbf{Motivation.}
Real driving records one outcome for each episode, while counterfactual evaluation requires
a matched outcome under an alternative action, which is never available in the real world. 
Therefore, we turn to the CARLA
simulator to build a benchmark in the controlled simulation \citep{carla}.
With CARLA, we can obtain counterfactual ground truths by running the same simulated world again.
For quantitative comparison, we also use CARLA to record a run of the same world where the event is never triggered, which serves as a reference video.

\noindent\textbf{Construction.}
In our benchmark, each case starts from one placement, an initial configuration of the
ego and one event agent, e.g., a
lead vehicle or a car waiting in a side street.
Both follow predefined open-loop motion scripts.
We then run the same
world three times, varying only the ego action and the presence of the
event, and record each run as an RGB sequence with synchronized ego motion.
The first run follows the executed trajectory and the event occurs, yielding
the factual log $(F,a_{\mathrm{obs}})$. The second follows the target trajectory
$a'$ while the rest of the world replays identically,
yielding the counterfactual ground truth $P$. The third follows the same target trajectory $a'$ in a run where the event
is never triggered, yielding the null reference $U$.
Each arm contains $25$ frames captured at $10$\,fps and $576\times320$
resolution. All
three arms share the $15$-frame history $H$ and diverge only in the
$10$-frame prediction window.
$P$ and $U$ are reserved for scoring.
Counterfactual edits retime the ego along its factual path. The ego
accelerates, slows down, or comes to a full stop, so the
query
changes only the ego motion.

\noindent\textbf{Composition.}
The benchmark contains $186$ cases drawn from $72$ placements across three towns
and three scenario types. The headline type, in which a vehicle emerges from a
side street, is the
cleanest test. Its event is first revealed in the factual continuation $F^{+}$,
which occupies the same time interval as the prediction window. The secondary clean type involves
a lead vehicle cutting in. The lead brake type is a confounded control, since an accelerating ego makes the
lead vehicle, which is already visible, loom larger,
mimicking the event signal through geometry alone. Cases from one placement
reuse the same initial setup and factual episode specification. Composition,
capture protocol, and benchmark checks are documented in
\cref{app:benchmark}.

\subsection{Metrics}
To evaluate a counterfactual prediction, we look at two dimensions. The
first is whether it depicts the right world, the one where the event
actually happened. The second is whether it depicts that world well,
with clear, seamless, and coherent content. We therefore score two
complementary axes, one semantic and one perceptual.

\noindent\textbf{Counterfactual signal recovered.}
The headline metric asks whether the prediction contains the right world. Let
$s(\cdot,\cdot)$ be the cosine similarity of corresponding frame embeddings,
averaged over the frames of the prediction window, and let
$\Delta(\hat{Y}) = s(\hat{Y},P)-s(\hat{Y},U)$ be the preference of a
prediction $\hat{Y}$ for the counterfactual over the null. To make this preference comparable across cases, we linearly rescale it using the two reference values as endpoints:
\begin{equation}
\mathrm{Rec}(\hat{Y}) \;=\;
\frac{\Delta(\hat{Y})-\Delta(U)}{\Delta(P)-\Delta(U)}.
\label{eq:rec}
\end{equation}
 We call this score the \emph{recovered fraction}. By construction,
  $\mathrm{Rec}(U)=0$ corresponds to complete event omission,
  $\mathrm{Rec}(P)=1$ corresponds to reproducing the reference event, and
  $\mathrm{Rec}(\hat{Y})=0.5$ means $\hat{Y}$ is equally similar to
  $P$ and $U$. A prediction whose preference lies between the two reference values scores in
  $(0,1)$, and the scale is not clipped outside them. For a set of cases, each $\Delta$ in
  \cref{eq:rec} is averaged over the cases.
For the encoders, we use DINOv2 ViT-B/14 \citep{dinov2} and
CLIP ViT-L/14 \citep{clip}, and write $\mathrm{Rec}_{\mathrm{D}}$ and
$\mathrm{Rec}_{\mathrm{C}}$ for the recovered fraction under each. 

\noindent\textbf{Perceptual fidelity.}
Because the true counterfactual $P$ exists, we measure quality as perceptual
distance
to it directly, using LPIPS \citep{lpips} between each predicted frame and the
corresponding frame of $P$. Unlike an embedding of the whole image,
LPIPS compares deep features \emph{spatially}, so locally wrong content, seams,
and blur all accrue distance.

\section{Experiments}
\subsection{Setup}

We compare our framework with the direct prediction on the same frozen backbone and case. Both are given
the same shared history $H$ and target trajectory $a'$. The direct prediction
$B$ follows the backbone's native history conditioning. Ours additionally uses
the factual continuation $F^{+}$ and its synchronized executed ego motion to
transport visual evidence into the counterfactual view.
The two procedures instantiate the two sides of \cref{eq:estimand}, so the
comparison exactly diagnoses the value of the factual evidence.
We evaluate two publicly
released models with different architectures.
Vista~\citep{vista} is a latent diffusion model conditioned on an anchor frame
and target trajectory. DrivingWorld~\citep{drivingworld} is an autoregressive
model over VQ tokens conditioned on frame history, pose, and heading. $B$ follows
each model's native conditioning protocol, and Ours retains that conditioning and
the frozen weights, adding the Abduce, Transport, Complete, and Combine stages. The main comparison reports means over five seeds: $B$ resamples the
native generation process, whereas Ours resamples completion while holding the
transported evidence fixed. We use the metrics of \cref{sec:benchmark}. Each inference
run uses one A100 GPU. 
Each value in the tables is the mean over five seeds, and the $\pm$ value beneath it is the maximum deviation from that mean.

\subsection{Qualitative Comparison}

\begin{figure}[t]
\centering
\includegraphics[width=\textwidth]{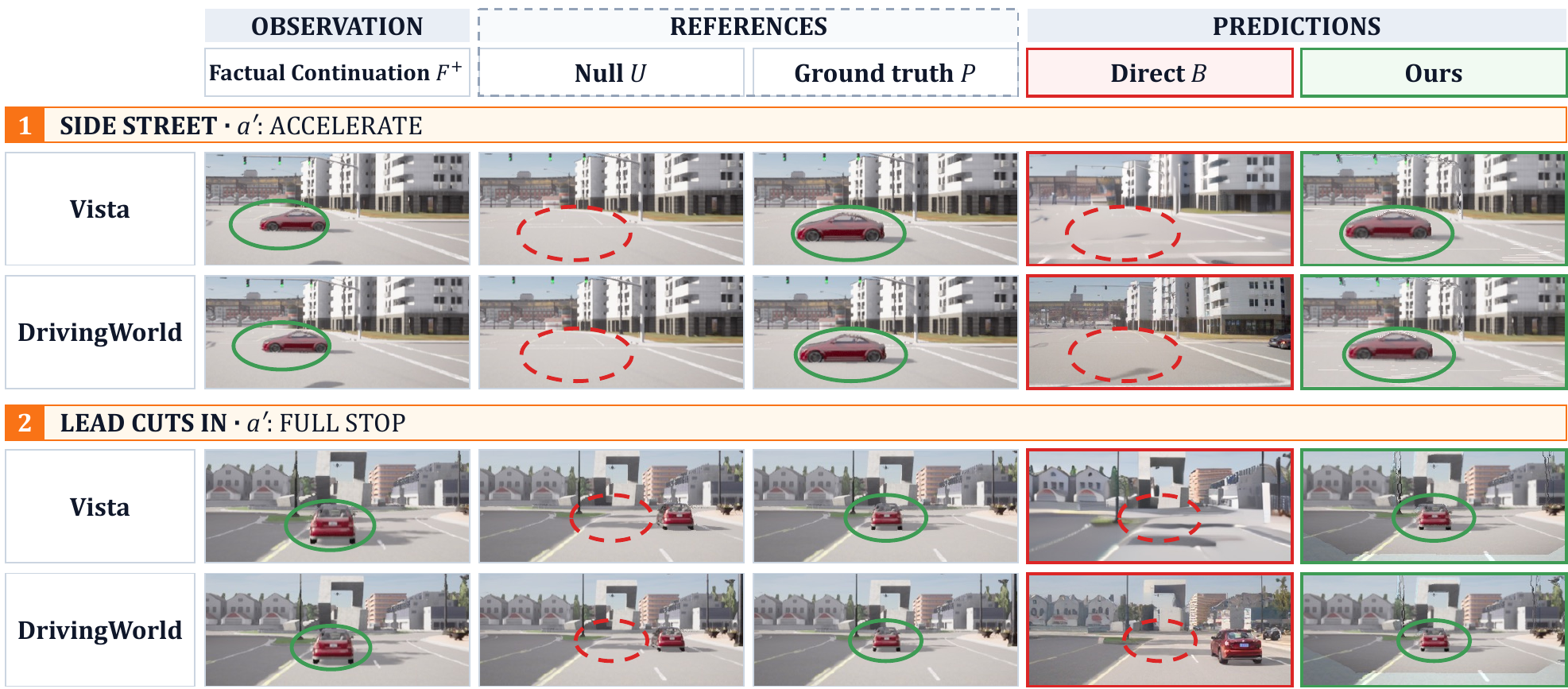}
\caption{Qualitative comparison at a representative frame late in the
prediction window. The upper block shows a vehicle emerging from a side
street under ego acceleration and the lower block a lead vehicle cutting in
while the ego brakes to a full stop. Each block has one row for Vista and one
for DrivingWorld. Columns show the factual continuation $F^{+}$, the
references used only for evaluation, $U$ (event-free null) and $P$
(counterfactual ground truth), and the predictions $B$ (direct prediction) and
Ours. Ellipses mark the event location, solid where the event vehicle is
present and dashed where it is absent. 
Within each row, $B$ and Ours use the same frozen model,
shared history $H$, and target trajectory $a'$.
$B$ omits or misplaces the event
vehicle, whereas Ours recovers it near the location shown in $P$.} 
\label{fig:qual}
\end{figure}

\cref{fig:qual} exposes a failure that visual quality alone would miss. Despite
producing fluent video, $B$ resembles $U$. Where a vehicle emerges from a side
street, neither backbone shows it. Where the lead vehicle cuts in, Vista
removes the vehicle and DrivingWorld leaves it in its original lane, so
neither preserves the realized event. In both scenarios Ours places the
vehicle at approximately the location and pose shown in $P$ on both
backbones.
Transport largely determines this geometry; the remaining seams and mild warp
artifacts are reflected in LPIPS.
\cref{fig:qualtime} in \cref{app:ablation} further shows one case across the full prediction window, showing the same pattern.

\subsection{Quantitative Comparison}

\cref{tab:delta} directly tests the two implications of the analysis.

\noindent\textbf{Direct predictions miss the realized event.}
For almost all scenarios, the mean for $B$ lies below $0.5$, ranging from $0.23$ to $0.45$, indicating $B$ remains closer to the
event-free null $U$ than to the matched counterfactual replay $P$, consistently
across both backbones and encoders. This pattern supports the analysis in
\cref{sec:analysis} that conditioning on the shared history and target trajectory
alone does not reliably preserve events specific to the episode that are revealed only in the
factual continuation $F^{+}$.

\noindent\textbf{Ours succeeds.}
With the same frozen backbones, Ours raises the overall recovered fraction to $0.64$--$0.70$. LPIPS falls from $0.423$ to $0.169$ on Vista and from $0.291$ to $0.211$ on DrivingWorld. Because the direct prediction and Ours use the same frozen backbones, these gains arise from supplying and preserving factual evidence specific to each episode, consistent with our theoretical analysis.

\begin{table}[t]
\centering
\small
\setlength{\tabcolsep}{2.1pt}
\begin{tabular}{l cc cc cc}
\toprule
& \multicolumn{2}{c}{$\mathrm{Rec}_{\mathrm{D}}{\uparrow}$} & \multicolumn{2}{c}{$\mathrm{Rec}_{\mathrm{C}}{\uparrow}$} & \multicolumn{2}{c}{LPIPS$\,{\downarrow}$} \\
\cmidrule(lr){2-3}\cmidrule(lr){4-5}\cmidrule(lr){6-7}
Scenario type & $B$ & Ours & $B$ & Ours & $B$ & Ours \\
\midrule
\multicolumn{7}{l}{\emph{Vista (diffusion)}} \\
side street & \seedcell{0.29}{.005} & \seedcell{\mathbf{0.75}}{.003} & \seedcell{0.25}{.007} & \seedcell{\mathbf{0.72}}{.001} & \seedcell{0.415}{.0075} & \seedcell{\mathbf{0.172}}{.0003} \\
lead cuts in & \seedcell{0.45}{.011} & \seedcell{\mathbf{0.73}}{.002} & \seedcell{0.38}{.056} & \seedcell{\mathbf{0.73}}{.005} & \seedcell{0.465}{.0046} & \seedcell{\mathbf{0.167}}{.0007} \\
lead brake & \seedcell{0.50}{.022} & \seedcell{\mathbf{0.59}}{.005} & \seedcell{0.41}{.036} & \seedcell{\mathbf{0.48}}{.002} & \seedcell{0.407}{.0029} & \seedcell{\mathbf{0.167}}{.0005} \\
\textbf{Overall} & \seedcell{0.38}{.006} & \seedcell{\mathbf{0.70}}{.002} & \seedcell{0.33}{.015} & \seedcell{\mathbf{0.65}}{.002} & \seedcell{0.423}{.0043} & \seedcell{\mathbf{0.169}}{.0003} \\
\midrule
\multicolumn{7}{l}{\emph{DrivingWorld (autoregressive)}} \\
side street & \seedcell{0.25}{.009} & \seedcell{\mathbf{0.74}}{.004} & \seedcell{0.23}{.010} & \seedcell{\mathbf{0.67}}{.003} & \seedcell{0.288}{.0014} & \seedcell{\mathbf{0.212}}{.0002} \\
lead cuts in & \seedcell{0.39}{.027} & \seedcell{\mathbf{0.67}}{.003} & \seedcell{0.28}{.012} & \seedcell{\mathbf{0.71}}{.004} & \seedcell{0.309}{.0048} & \seedcell{\mathbf{0.214}}{.0003} \\
lead brake & \seedcell{0.37}{.018} & \seedcell{\mathbf{0.55}}{.003} & \seedcell{0.23}{.016} & \seedcell{\mathbf{0.51}}{.006} & \seedcell{0.284}{.0021} & \seedcell{\mathbf{0.208}}{.0003} \\
\textbf{Overall} & \seedcell{0.31}{.008} & \seedcell{\mathbf{0.67}}{.001} & \seedcell{0.24}{.007} & \seedcell{\mathbf{0.64}}{.001} & \seedcell{0.291}{.0013} & \seedcell{\mathbf{0.211}}{.0002} \\
\bottomrule
\end{tabular}
\caption{Main results by scenario type. Each pair of columns compares the
direct prediction $B$ with Ours on the same frozen backbone, and the better
value is in bold. $\mathrm{Rec}_{\mathrm{D}}$ and $\mathrm{Rec}_{\mathrm{C}}$ are recovered
fractions computed with DINOv2 and CLIP, and LPIPS is computed against $P$.}
\label{tab:delta}
\end{table}
\subsection{Ablations}

\begin{table}[t]
\centering
\small
\setlength{\tabcolsep}{1.25pt}
\begin{tabular}{cccc ccc ccc}
\toprule
\multicolumn{4}{c}{\textbf{Components}} & \multicolumn{3}{c}{\textbf{Vista}} & \multicolumn{3}{c}{\textbf{DrivingWorld}} \\
\cmidrule(lr){1-4}\cmidrule(lr){5-7}\cmidrule(lr){8-10}
Tr & MF & Cm & Cb &
$\mathrm{Rec}_{\mathrm{D}}{\uparrow}$ & $\mathrm{Rec}_{\mathrm{C}}{\uparrow}$ & LPIPS$\,{\downarrow}$ &
$\mathrm{Rec}_{\mathrm{D}}{\uparrow}$ & $\mathrm{Rec}_{\mathrm{C}}{\uparrow}$ & LPIPS$\,{\downarrow}$ \\
\midrule
-- & -- & -- & --
  & \seedcell{0.38}{.006} & \seedcell{0.33}{.015} & \seedcell{0.423}{.0043}
  & \seedcell{0.31}{.008} & \seedcell{0.24}{.007} & \seedcell{0.291}{.0013} \\
$\checkmark$ & $\checkmark$ & -- & --
  & \seedcell{0.68}{.003} & \seedcell{0.65}{.002} & \seedcell{0.195}{.0002}
  & \seedcell{0.67}{.001} & \seedcell{0.66}{.001} & \seedcell{0.238}{.0002} \\
$\checkmark$ & -- & $\checkmark$ & $\checkmark$
  & \seedcell{0.69}{.005} & \seedcell{0.64}{.003} & \seedcell{0.187}{.0004}
  & \seedcell{0.65}{.003} & \seedcell{0.63}{.003} & \seedcell{0.223}{.0005} \\
$\checkmark$ & $\checkmark$ & $\checkmark$ & $\checkmark$
  & \seedcell{0.70}{.002} & \seedcell{0.65}{.002} & \seedcell{0.169}{.0003}
  & \seedcell{0.67}{.001} & \seedcell{0.64}{.001} & \seedcell{0.211}{.0002} \\
\bottomrule
\end{tabular}
\caption{Ablation of transport (Tr), filling from multiple frames (MF), completion
(Cm), and the Combine stage (Cb). The first row is the direct prediction $B$,
and the last row is the full method.}
\label{tab:ablation}
\end{table}

Following the pipeline described in \cref{sec:method}, \cref{tab:ablation} isolates four implementation components: transport (Tr), its refinement that fills from multiple frames (MF), completion (Cm), and the Combine stage
(Cb). Tr moves evidence from the corresponding factual frame into the
counterfactual view, MF uses nearby factual frames to fill residual holes, Cm
fills and harmonizes the remaining unsupported regions with the frozen world
model, and Cb restores the transported evidence after completion. We compare
the direct prediction $B$; Tr+MF, which retains pixels from $B$ in the
unsupported regions; Tr+Cm+Cb, which uses the time-aligned factual frame
followed by the Complete and Combine stages; and the full Tr+MF+Cm+Cb pipeline.

 \noindent\textbf{Transport carries the event signal.}
The Tr+MF variant reaches
$\mathrm{Rec}_{\mathrm{D}}$/$\mathrm{Rec}_{\mathrm{C}}$ of $0.68/0.65$ on Vista and
$0.67/0.66$ on DrivingWorld, compared with $0.38/0.33$ and $0.31/0.24$ for
the direct prediction. These results show that transport recovers
much of the event signal lost by the direct prediction.

\noindent\textbf{The Complete and Combine stages improve visual fidelity.}
Transport with filling from multiple frames recovers the event but can still leave visible seams and unsupported regions inherited from the direct prediction.
Adding the Complete and Combine stages reduces LPIPS from $0.195$ to $0.169$ on Vista and from $0.238$ to $0.211$ on DrivingWorld.
The lower LPIPS values are consistent with fewer visual artifacts after these stages.
The full version with filling from multiple frames also achieves a higher
recovered fraction and lower LPIPS than the variant using one frame, suggesting that
agreement across frames produces cleaner and more reliable transported evidence.
\cref{app:ablation} reports intermediate outputs after completion and examines
how the choice of factual evidence affects transport.

\subsection{Cost}
On a single A100 GPU, Ours takes about $90$\,s per Vista case and $108$\,s per DrivingWorld case, compared with $47$\,s and $45$\,s for direct prediction.
Our method adds inference-time processing while keeping all model parameters
frozen. The resulting runtime is practical for the offline
counterfactual analysis considered here. Runtimes for each stage and peak memory
usage are reported in \cref{app:method}.

\section{Conclusion}
A common practice in driving world models is to treat direct action-conditioned
prediction as counterfactual prediction. In this paper, we first identified a
fundamental mismatch between direct prediction and counterfactual prediction. The direct prediction conditions on the shared history and
target trajectory, but not on the factual continuation already observed when the counterfactual query is posed. It therefore
marginalizes events that should remain unchanged under the counterfactual action
rather than preserving them. To make this failure measurable, we constructed a
controlled benchmark with factual outcomes and matched counterfactual ground truths. We further built a deliberately simple pipeline that transports evidence, as
a constructive check that supplying the missing evidence closes much of the
gap. Across two world models, it recovered much of the event signal
lost by direct predictions and improved visual fidelity using frozen model
weights. 
For limitations and future work, please refer to \cref{app:limitations}.

\bibliographystyle{plainnat}
\bibliography{references}

\newpage
\appendix
\begin{appendices}
\crefalias{section}{appendix}
\crefalias{subsection}{appendix}
\startcontents[appendices]
\printcontents[appendices]{l}{1}{\setcounter{tocdepth}{2}}

\section{Benchmark Details}
\label{app:benchmark}

This section documents the construction and scoring of the $186$ benchmark
cases.

\subsection{Composition}

\cref{tab:appcomp} breaks the $186$ cases down by scenario type and target
ego action. The cases come from the $72$ placements of \cref{sec:benchmark},
with $27$ for lead brake, $26$ for side street, and $19$ for lead cuts in. Every placement contributes an acceleration case, and
the braking and full stop cases cover subsets of these placements. The town
distribution is Town01 ($60$), Town03 ($72$), and Town10HD ($54$). We also
collected $10$ pedestrian crossing cases. This sample is too small for a
separate comparison, so the reported results use the $186$ vehicle cases.

\subsection{Capture and Sequence Construction}

All data are rendered in CARLA~0.9.15 in synchronous mode with a fixed
simulation step of $0.05$\,s. A single front RGB camera ($576\times320$,
$70^\circ$ horizontal field of view) is mounted
$1.5$\,m forward of and $1.5$\,m above the ego actor origin of a
\texttt{vehicle.tesla.model3}. Two simulation steps are taken between stored frames.

Frames $0$--$14$ form the shared history $H$, and frames $15$--$24$ form the
prediction window. In $F$, the prediction window is the factual continuation
$F^{+}$. The replays $P$ and $U$ use the same scene setup during the history,
apart from small rendering variations. In $U$, the event vehicle is present
with the same starting state, and its scripted maneuver is pushed beyond the
captured window. The lead vehicle keeps its speed, the side street vehicle
stays waiting, and the cutting-in vehicle keeps its lane. From the first
prediction frame, the target trajectory scales the displacement between
consecutive factual positions by $1.6$, $0.4$, or $0$ for acceleration,
braking, or a full stop.

\subsection{Metadata for Each Case}

Each case includes a \texttt{meta.json} file. It records the case identifiers,
map, action edit, ego and event vehicle spawn poses, and the parameters of the
scripted vehicle motion. It also stores the ego positions and headings for
$F$, $P$, and $U$, together with a vehicle visibility summary and minimum
approach distance.
Some side street cases additionally record the distance to the junction.
These records provide the camera motion used by evidence transport and the
geometry used by the benchmark checks.

\subsection{Benchmark Checks}

The geometric check determines inclusion in the benchmark. Using the recorded
ego and vehicle trajectories, a forward $70^\circ$ view, and a maximum
distance of $60$\,m, it verifies that the scripted event vehicle is visible in
the factual camera view in at least two frames. All $186$ reported cases pass this check. A separate image check
compares $P$ and $U$. It checks their agreement at the end of the shared
history and the appearance of a localized difference during the prediction
window. This second check is an audit after capture and uses simple image
thresholds. It passes $167$ cases and flags $19$ for review. The flags concentrate in
combinations where the action edit weakens the visible event, for example a
braking edit that lets the lead vehicle recede, and in cases whose image
differences are close to rendering noise. Scoring without the flagged cases
leaves the comparison between the direct prediction and Ours unchanged, so all
$186$ cases remain in the reported benchmark. 

\subsection{Scoring Metrics}

For a prediction $\hat{Y}$, the preference $\Delta(\hat{Y})$ of
\cref{sec:benchmark} is computed as
\[
\Delta(\hat{Y}) = \frac{1}{10}\sum_{t=15}^{24}
   \big[\cos(\phi(\hat{Y}_t),\phi(P_t)) - \cos(\phi(\hat{Y}_t),\phi(U_t))\big],
\]
where $\phi$ is the frozen encoder and its output is normalized to unit $L_2$
norm.

For each case, let
\[
d=\frac{1}{10}\sum_{t=15}^{24}
\big[1-\cos(\phi(P_t),\phi(U_t))\big].
\]
Then $\Delta(P)=d$ and $\Delta(U)=-d$, so the recovered fraction in
\cref{eq:rec} becomes $(\Delta(\hat{Y})+d)/(2d)$. For a set of cases, we
average $\Delta$ and $d$ separately and report
$(\overline{\Delta}+\bar d)/(2\bar d)$. Scores outside $[0,1]$ are retained.
LPIPS is computed with the AlexNet backbone and averaged over the ten frames
of the prediction window. For DrivingWorld, the reference frames are resized
to the model's $512\times256$ output size before comparison.

\begin{table}[t]
\centering
\small
\setlength{\tabcolsep}{2.0pt}
\begin{tabular}{lrrrr}
\toprule
Scenario type & Total & Accel. & Brake & Full stop \\
\midrule
side street & $60$  & $26$ & $17$ & $17$ \\
lead cuts in & $45$  & $19$ & $13$ & $13$ \\
lead brake & $81$  & $27$ & $27$ & $27$ \\
\midrule
Total                         & $186$ & $72$ & $57$ & $57$ \\
\bottomrule
\end{tabular}
\caption{Benchmark composition by scenario type and target ego action.}
\label{tab:appcomp}
\end{table}
\section{Method Details}
\label{app:method}

\noindent\textbf{Input resolution.}
Vista uses the benchmark frames at their native resolution of
$576\times320$. For DrivingWorld, we resize each frame to $512\times284$ with
bicubic interpolation and crop $14$ pixels from the top and bottom, producing
the model's $512\times256$ input while preserving the aspect ratio. All
constants in this section were set during initial development and then kept
fixed for the reported runs.

\noindent\textbf{Abduce and Transport.}
We estimate relative depth with Depth Anything V2 Small
\citep{depthanything}. The fixed image resolution and horizontal field of view
determine the focal length and image center. We use a road patch near the bottom center of the image and a camera height
constant of $1.8$\,m above the road surface to convert relative depth into
distance. The $1.5$\,m mount height in \cref{app:benchmark} is measured from the ego
actor origin rather than the road.

We remove pixels at sharp depth changes, where the estimated 3D position is
less reliable, using a relative depth gradient threshold of $0.15$. We sample
the source image at twice its resolution and keep the nearest projected 3D
point at each target pixel. Three rounds of neighboring pixel averaging close
narrow holes of $1$--$3$ pixels.

In filling from multiple frames (MF), a pixel with one available projection
is retained. When several frames project to the same pixel and their average spread across
channels is below $28$ intensity levels, we retain their median RGB value.
Otherwise the pixel stays unsupported and is left to the Complete stage. Before filling a region,
we match the donor frame's brightness and contrast to the accepted evidence
near its boundary.

\noindent\textbf{Complete.}
On Vista, the native EDM diffusion sampler uses $25$ steps and starts at
schedule index $14$ ($\sigma\approx6.4$). The $15$ history frames remain fixed.
We first resize the evidence mask to the latent representation. Each
$8\times8$ cell stores the fraction of its pixels covered by evidence. This
fraction serves as $M$ in the restoration step, so a partly covered cell is
only partly fixed to the evidence. On
DrivingWorld, an evidence token is kept fixed when transported
evidence covers at least $60\%$ of its $16\times16$ image patch.

\noindent\textbf{Combine.}
To form the region where $\alpha_t=1$, we shrink the support mask $M_t$ by
$2$ pixels, which removes uncertain boundary pixels. The transition to the completed image is blended over $12$ pixels for
Vista and $24$ pixels for DrivingWorld. The wider transition reduces visible
boundaries between DrivingWorld's $16\times16$ image tokens. The map $\mathrm{cc}$ applies a linear color adjustment that matches the
model output to the transported evidence near the boundary. For each channel, the contrast scale lies in $[0.8,1.25]$ and the
brightness shift lies within $\pm25$ intensity levels. We average the
adjustment with that of the previous frame using weight $0.5$ to keep the
appearance stable over time.

\noindent\textbf{Cost breakdown.}
Timings begin after model loading, use one A100 with batch size $1$, and are
averaged over three cases, one from each scenario type. Ours computes the
direct prediction $B$ once, at about $47$\,s for Vista and $45$\,s for
DrivingWorld, since $B$ fills the unsupported regions of the input video for
Complete. Beyond this, depth estimation takes about $3$\,s, transport on the
CPU takes about $18$\,s, and Combine takes about $2$\,s for both models.
Completion takes about $21$\,s for Vista and $40$\,s for DrivingWorld. Up to
rounding, these components sum to the total runtimes of about $90$\,s and
$108$\,s reported in the main text. Peak GPU memory is $49$\,GB for Ours and
$39$\,GB for $B$ on Vista. Both methods use about $12$\,GB on
DrivingWorld. Experiments run on Ubuntu 24.04, with PyTorch 2.0.1 and CUDA
11.8 for Vista and PyTorch 2.5.1 and CUDA 12.1 for DrivingWorld.

\section{Additional Analyses}
\label{app:ablation}

\subsection{The Recovered Event Develops over Time}

\begin{figure}[t]
\centering
\includegraphics[width=\columnwidth]{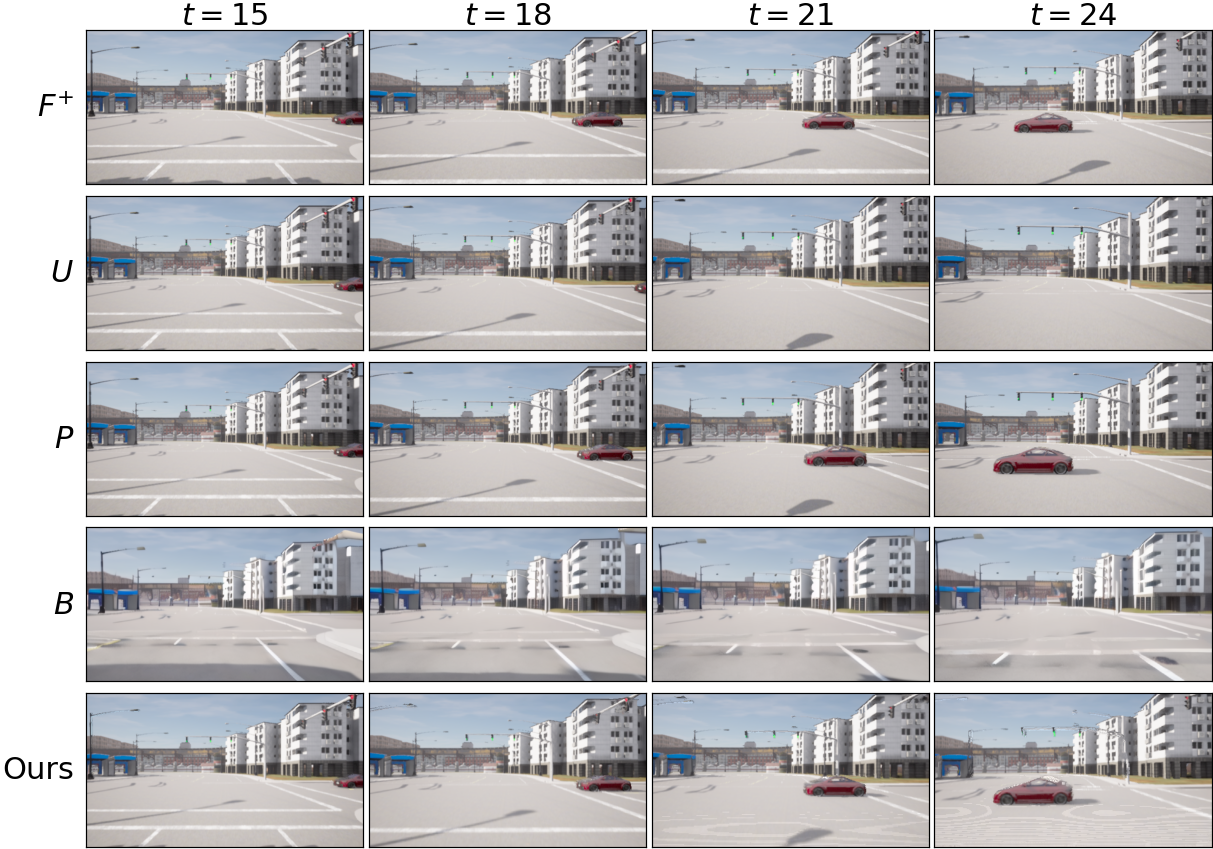}
\caption{A temporal comparison for a side street case in Town03 with
Vista. The columns show frames $15$, $18$, $21$, and $24$. The rows show the
factual continuation $F^{+}$, null reference $U$, counterfactual ground truth
$P$, direct prediction $B$, and Ours. The vehicle emerges over time in
$F^{+}$, $P$, and Ours.}
\label{fig:qualtime}
\end{figure}

\cref{fig:qual} in the main text compares the methods at one late frame.
The prediction is a video, so we also follow one case across the whole
prediction window. In \cref{fig:qualtime}, a side street case with Vista,
the vehicle becomes visible in Ours by frame $18$ and advances across the
junction as it does in $P$. The $B$ frames grow blurrier over the window,
while Ours keeps the background sharp. The recovered event thus develops over
time rather than appearing at a single frame.

\FloatBarrier
\FloatBarrier

\subsection{Results Are Stable across Seeds}
For the seed means reported in \cref{tab:delta,tab:ablation}, we first
average each case's $\Delta$ across seeds and then aggregate across cases.
Vista uses a deterministic per-case seed and DrivingWorld a global seed, each
varied across the five runs.
Across the five seeds, the overall DINOv2 recovered fraction of $B$ spans
$0.377$--$0.388$ on Vista and $0.305$--$0.321$ on DrivingWorld, and overall
LPIPS spans $0.419$--$0.426$ and $0.290$--$0.292$. The low scores of the
direct prediction are therefore stable across seeds, and the deviations of
Ours are smaller still, as the $\pm$ values in the two tables show. $P$ and $U$ are most similar for lead brake cases, so the denominator of the recovered fraction is small there and small changes in the encoder output are magnified.

\subsection{Complete Fills the Holes and Combine Restores the Evidence}

\cref{tab:ablation} compares end-to-end variants of the pipeline. Such a
comparison leaves open what the last two stages contribute individually and
why the Combine stage is needed. We therefore score the same runs at three
checkpoints, before completion, after completion, and after the final Combine
stage. \cref{tab:stages} reports the checkpoints as means over five seeds; the intermediate stage deviates from its mean by at
most $0.003$ in the recovered fractions and $0.001$ in LPIPS. On DrivingWorld, both
recovered fractions dip after completion and return after Combine, and LPIPS
follows the same pattern. The dip reflects the loss from passing the full
image through the token encoder and decoder, and the recovery quantifies the
effect of the Combine stage. On Vista, whose completion operates in a
continuous latent space, the dip nearly vanishes, and LPIPS improves at each
stage. In sum, Complete supplies the unsupported regions, Combine keeps the
transported evidence intact, and the two together lower LPIPS on both models.

\begin{table}[t]
\centering
\small
\setlength{\tabcolsep}{2.8pt}
\begin{tabular}{l ccc ccc}
\toprule
& \multicolumn{3}{c}{\textbf{Vista}} & \multicolumn{3}{c}{\textbf{DrivingWorld}} \\
\cmidrule(lr){2-4}\cmidrule(lr){5-7}
Stage & $\mathrm{Rec}_{\mathrm{D}}$ & $\mathrm{Rec}_{\mathrm{C}}$ & LPIPS & $\mathrm{Rec}_{\mathrm{D}}$ & $\mathrm{Rec}_{\mathrm{C}}$ & LPIPS \\
\midrule
Tr+MF & $0.68$ & $0.65$ & $0.195$ & $0.67$ & $0.66$ & $0.238$ \\
$+$ Complete & $0.69$ & $0.62$ & $0.180$ & $0.52$ & $0.45$ & $0.261$ \\
$+$ Combine (full) & $0.70$ & $0.65$ & $0.169$ & $0.67$ & $0.64$ & $0.211$ \\
\bottomrule
\end{tabular}
\caption{Scores at the three checkpoints of the pipeline, as means over five
seeds. Tr+MF is transport with filling from multiple frames, before the frozen
model is used. Higher recovered fractions and lower LPIPS are better. The
first and last rows match the corresponding rows of \cref{tab:ablation}.}
\label{tab:stages}
\end{table}

\subsection{Transport Requires the Correct Episode and Time}

The main comparison shows that Ours recovers the event signal, and this
section asks where the gain comes from. If pasting any additional pixels
helped, evidence from a wrong time or a wrong episode would score as well as
the matching evidence. This diagnostic therefore fixes the pipeline and varies
only the evidence source, using one seed and transport from a single frame.
We first use the factual frame at the matching time. We then replace it with the factual frame five frames earlier, with the
final history frame $F_{14}$, or with a frame from another case with the same
scenario and action edit. The donor from another case is selected
from the same town when possible and has a similar minimum approach distance.

\FloatBarrier
\begin{table}[t]
\centering
\small
\setlength{\tabcolsep}{2.8pt}
\begin{tabular}{l ccc ccc}
\toprule
& \multicolumn{3}{c}{\textbf{Vista}} & \multicolumn{3}{c}{\textbf{DrivingWorld}} \\
\cmidrule(lr){2-4}\cmidrule(lr){5-7}
Evidence & $\mathrm{Rec}_{\mathrm{D}}$ & $\mathrm{Rec}_{\mathrm{C}}$ & LPIPS & $\mathrm{Rec}_{\mathrm{D}}$ & $\mathrm{Rec}_{\mathrm{C}}$ & LPIPS \\
\midrule
matching evidence   & $0.66$ & $0.65$ & $0.225$ & $0.67$ & $0.64$ & $0.261$ \\
five frames earlier & $0.40$ & $0.44$ & $0.276$ & $0.41$ & $0.47$ & $0.323$ \\
final history frame  & $0.35$ & $0.39$ & $0.229$ & $0.36$ & $0.43$ & $0.290$ \\
different case            & $0.62$ & $0.58$ & $0.564$ & $0.64$ & $0.59$ & $0.556$ \\
\midrule
direct prediction $B$ & $0.38$ & $0.32$ & $0.419$ & $0.31$ & $0.24$ & $0.291$ \\
\bottomrule
\end{tabular}
\caption{Evidence source controls on $186$ cases using one seed. Each
evidence row uses transport from one frame, with the corresponding direct
prediction $B$ supplying the remaining pixels. Higher recovered fractions and
lower LPIPS are better.}
\label{tab:placebo}
\end{table}

As \cref{tab:placebo} shows, evidence from the matching time and case
reaches a $\mathrm{Rec}_{\mathrm{D}}$ of $0.66$ on Vista and $0.67$ on DrivingWorld.
Evidence from an earlier time produces $0.35$--$0.41$, close to the direct
prediction at $0.31$--$0.38$. Evidence from another case still contains a
similar vehicle event and therefore reaches $0.62$--$0.64$, but its LPIPS
rises to $0.564$ on Vista and $0.556$ on DrivingWorld. $\mathrm{Rec}_{\mathrm{C}}$
follows the same pattern. Together, the two metrics
show that successful transport depends on evidence from the correct episode
and time.

\section{Limitations and Future Work}
\label{app:limitations}

The choices that enable our controlled open-loop setting also introduce
several directions for future work. The benchmark scripts the surrounding
agents, which is what makes the matched counterfactual reference $P$
obtainable. Over longer horizons, surrounding agents react to the ego, and
transported evidence then preserves behavior that the counterfactual action
would have changed. For example, a pedestrian who would have stopped if the
ego had slowed would keep walking in the transported evidence. A natural next
step is to detect such cases, for example with posterior predictive checks on
the abduced world, and to extend the protocol to closed-loop scenario suites
with reactive agents \citep{bench2drive,nuplan}.

On the experimental side, both world models are evaluated outside their
training render domain. The causal analysis does not depend on the render
domain, and each model is evaluated under its authors' released inference
protocol, yet a replication with a world model trained on the benchmark's
render domain would complement the present comparison. On the method side,
monocular depth error degrades the transported evidence and can leave visible
seams, and the unsupported holes inherit the completion quality of the
backbone. Since every network in the pipeline is frozen and replaceable,
advances in depth estimation and in world models transfer to the method
directly.

Finally, the method reads the factual continuation, which exists only after
the episode has been recorded, so it serves the retrospective queries of
\cref{subsec:setup}. The open-loop protocol is a controlled research
instrument, so applications such as liability assessment would additionally
call for the reactive extensions discussed above. Extending the framework to
decision time, where the
outcome is still unobserved, is a further direction.

\clearpage

\end{appendices}

\end{document}